\documentclass[10pt,twocolumn,letterpaper]{article}

\usepackage{wacv}
\usepackage{times}
\usepackage{epsfig}
\usepackage{graphicx}
\usepackage{amsmath}
\usepackage{amssymb}

\usepackage[T1]{fontenc}
\usepackage[utf8]{inputenc}
\usepackage{authblk}

\usepackage[bookmarks=false]{hyperref}

\wacvfinalcopy % *** Uncomment this line for the final submission

\def\wacvPaperID{****} % *** Enter the wacv Paper ID here

\ifwacvfinal\fi
\begin{document}

%%%%%%%%% TITLE
\title{Image2GIF: Generating Cinemagraphs using Recurrent Deep Q-Networks}

% Authors at the same institution
%\author{First Author \hspace{2cm} Second Author \\
% Authors at different institutions
%\author{First Author \\
%\and
%Second Author \\

%\makeatletter
%\renewcommand\AB@affilsepx{, \protect\Affilfont}
%\makeatother

\author[1]{Yipin Zhou \thanks{This work was done while the author was an intern at Yahoo Research.}}
\author[2]{Yale Song}
\author[1]{Tamara L. Berg}
\affil[1]{University of North Carolina at Chapel Hill} 
\affil[2]{Yahoo Research}
\affil[ ]
{\small\url{http://bvision11.cs.unc.edu/bigpen/yipin/WACV2018}}

\maketitle
\ifwacvfinal\thispagestyle{empty}\fi

%%%%%%%%% ABSTRACT
\begin{abstract}
Given a still photograph, one can imagine how dynamic objects might move against a static background. This idea has been actualized in the form of cinemagraphs, where the motion of particular objects within a still image is repeated, giving the viewer a sense of animation. In this paper, we learn computational models that can generate cinemagraph sequences automatically given a single image. To generate cinemagraphs, we explore combining generative models with a recurrent neural network and deep Q-networks to enhance the power of sequence generation. To enable and evaluate these models we make use of two datasets, one synthetically generated and the other containing real video generated cinemagraphs. Both qualitative and quantitative evaluations demonstrate the effectiveness of our models on the synthetic and real datasets.
\end{abstract}

%%%%%%%%% BODY TEXT
\section{Introduction}

\begin{figure}[t]
\begin{center}
\includegraphics[width=1\linewidth]{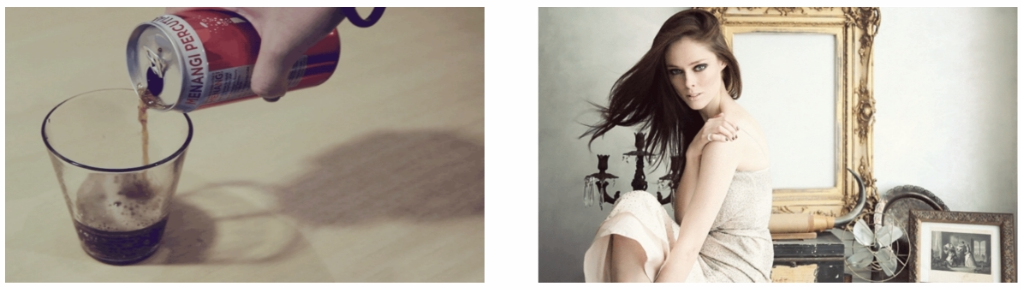}
\end{center}
   \caption{Example images where one can easily imagine dynamic motion -- the coke will flow, the hair will wave.}
\label{fig:intro}
\end{figure}

Based on our life-long observations of the natural world, humans have the ability to reason about visual appearances of static and dynamic objects. In particular, given an image, one can easily picture which objects will move and how they might move in the future. For instance, given an image showing a can pouring liquid into a cup, as illustrated in Fig. \ref{fig:intro} (left), one can imagine how the liquid must be flowing due to gravity and what that might look like. Or, given an image showing a woman with billowing hair (right), we can imagine how her hair might wave in the breeze.

Cinemagraphs are still photographs in which a repeated movement of one or more objects occur, often served as an animated GIF to produce a repeating effect. Often these objects are natural entities, e.g. water flowing in a fountain, plants blowing in the breeze against a still background, and the steps of an escalator ascending. In this way, cinemagraphs can be seen as highlighting and illustrating the motion of specific dynamic objects in static scenes. Because of this, we posit that understanding and generating cinemagraphs is useful for providing insight into the world -- for understanding what objects can move and how they move -- and for evaluating how well computational models learn to represent the dynamic nature of objects. 

In this work, our goal is to learn a computational model to reason about visual appearances of static and dynamic objects. Specifically, we train a generative model to animate part of the input photograph and output a cinemagraph sequence. Recent generation works have explored the tasks of generating novel images~\cite{GANs,DRAW,laplacian,DCGAN,Style_gen,SSIM,energy,Wasserstein_GAN,pix2pix2016,CycleGAN2017,BEGAN,progressive_GAN} and future video frames~\cite{LSTM,Ranzato2014,Yann,temp,visual_dynamics,VPN,Liang_2017_ICCV,Hierarchical_future,Unsuper_physical,jacob_iccv2017}, but this work is the first to explore the generation of cinemagraphs from a single image, which include the features of both images (static) and videos (dynamic). Cinemagraph generation from a single image is very challenging because the model needs to figure out both where to animate (localization) and how to animate (generation). 

With both dynamic and static characteristics, cinemagraphs highlight particular
objects within an image, bestowing a sense of feeling or emphasis to a static image. As a result, they have become quite popular on the Internet (querying cinemagraph on Google yields 1,960,000 results). However, one drawback is that creating a cinemagraph requires tedious manual work: people use videos to create a cinemagraph using commercial softwares, fixing part of frames and animating others. Thus, one useful practical benefit of this work is to automate the process of generating a cinemagraph from a single image. Other benefits include learning about what objects can move in a scene and how they move. This can be accomplished by developing models to generate cinemagraphs and then evaluating how well their predictions match the reality.

We explore different model architectures stacked in a recurrent structure to recursively generate cinemagraph sequences. Additionally, we incorporate Deep Q-networks, a deep reinforcement learning algorithm (Sec.~\ref{methods}), to improve performance by brining in stochasticity to the model. To enable evaluation of these models, we use two new datasets: one is synthetically generated, the other is cinemagraph data we collected from the Internet (Sec.~\ref{datasets}).Finally, we provide both quantitative and qualitative evaluation of the models on our synthetic and real datasets (Sec.~\ref{evaluation}).
%\textcolor{green}{The real cinemagraph dataset is very challenging, for instance, in the water flowing category, the fixed background can be varied and the water can flow in difference directions (e.g. horizontally or vertically). This requires models to learn the semantics of the moving objects, their appearance attributes, and how to animate them while also learning what image content should remain static.} 

Our contributions include: 1) We introduce a new task of generating cinemagraphs from a single image, 2) We release two new datasets for our task, 3) We explore and evaluation various recurrent generative model structures combined with deep Q-net for cinemagraph generation.

%-------------------------------------------------------------------
\section{Related Work}
\noindent {\bf Image generation:} Applying generative models on natural images has attracted a great deal of attention in recent years. Gregor~\etal~\cite{DRAW} recurrently generate different areas of a single image using an attention mechanism with variational autoencoders. Goodfellow~\etal~\cite{GANs} proposed generative adversarial networks (GANs) that greatly improve image generation quality. GANs consist of a generative model and a discriminative model, trained jointly for enhancing the realism of generated images. Radford~\etal~\cite{DCGAN} explored combining deep convolutional neural networks in GANs to further improve the image quality. Pathak~\etal~\cite{CVPR16context} proposed a network to generate contents of an arbitrary image region conditioned on its surroundings. Finally, Zhao~\etal~\cite{energy} proposed energy-based GANs that consider the discriminative model as an energy function. Following these great successes, we also explore the use of GANs for cinamagraph sequence generation.

\noindent {\bf Video frame generation:} As a step beyond creating realistic images, video generation requires the model to incorporate temporal information into the generation process. Therefore, video generation models usually generate frames conditioned on the previous frame (or several previous frames), rather than generating from noise, as is usually done in image generation tasks. Some related approaches \cite{LSTM,Ranzato2014} train generative models either to reconstruct input video frames or to generate the next few consecutive frames, learning representations in an unsupervised framework. Mathieu~\etal~\cite{Yann} combines a mean square error (MSE) loss with an adversarial loss to produce high-quality generation results. Vondrick~\etal~\cite{scene_dynamics} proposed a GAN with a spatio-temporal convolutional architecture to learn a scene's foreground and background simultaneously. Zhou~\etal~\cite{temp} incorporated a recurrent neural network with a generative model to generate frames showing future object states in timelapse videos. Inspired by their work, we apply an LSTM~\cite{LSTM1997} to encode temporal information during cinemagraph generation.

\noindent {\bf Deep reinforcement learning:} Reinforcement Learning (RL) has been applied to many applications~\cite{obstacle,helicopter,trajectory}. Recently, works combining RL (especially model-free RL) with deep neural networks have attracted extensive attention~\cite{ATARI,continuous,navi,Asy,action_atari}. Model-free RL algorithms can be divided into two types: Q-learning and policy gradient learning. Deep Q-learning (DQL) predicts which action to take at each time step to maximize future rewards, while policy gradient methods directly optimize a policy of expected reward using gradient descent. For example, Mnih~\etal~\cite{ATARI} use DQL to control an agent to play ATARI games (manipulating the joystick), while Gu~\etal~\cite{q_continue} proposed a DQL method with a continuous action space. Silver~\etal~\cite{deter}, Lillicrap~\etal~\cite{continuous}, Mnih~\etal~\cite{Asy}, and Schulman~\etal~\cite{highdim} make use of an actor-critic or an asynchronous variant of the actor-critic algorithm based on deterministic policy gradient that can solve the tasks with continuous action space, e.g., cartpole swing-up, 3D locomotion, or other robotics tasks. 

On the Atari task, Mnih~\etal~\cite{ATARI} utilized a Deep Q-network (DQN) to select an action $a$, which they apply to the state $s$ in order to achieve maximal award for each time step. Here, $a$ (e.g. joystick operation), $s$ (game screen) and award (game score) can be easily defined. Moreover the action space is discrete (e.g. agents can only move up and down). Later work~\cite{q_continue} applies a DQN to a continuous action space for controlling robotic arms. Those works have clear definitions for action space (e.g., angles, positions). Previous success of Deep Q-learning witnessed the potential of DQN for sequential decision-making. In this work, we apply a DQN in a sequence generation task, which differs with previous works in that our action space consists of abstract concepts that help with the sequence decision making. In other words, we select actions to apply to the current image to further lower the generative loss (maximize award).

%In this work, we want to make the decision of what 'action' to take to generate future frames.  We hypothesize that incorporating DRL structure may help with this goal due to its potential for decision-making for future steps. However, in traditional reinforcement learning tasks, the action space (no matter whether discrete or continuous) is generally meaningful and nameable. For example, actions might be represented as moving in a particular direction, distance, or angle, etc. The states after applying an action to the current state are also usually easily described as a location, position, etc. 

%In our generation task, one challenge is to define the explicit meaning of taking an 'action' and to model what the state will look like after the action has been applied. To accomplish this, we quantize the action space into a discrete space and pass it through the decoding part of the generative model, to decode how the quantized action will effect the future state (future frame).Regarding choice of policy gradient vs. deep Q-Learning, in policy gradient methods, the policy needs to be optimized end-to-end, which is not straightforward to tackle with the generative model. Therefore, we make use of a deep Q-learning that can explicitly output actions for states in our generative model. %For more details please refer Sec \ref{sec:recurrent q-net}.

%-------------------------------------------------------------------
\section{Methods}
\label{methods}
Our goal is to generate a cinemagraph with $n$ frames $\mathrm{Y} = \{y_1, ..., y_n\}$ given a static image $x$ as input. This requires building models that can generate depictions of future states of moving objects over time. Given an image, we expect the model to learn both what part of the image should be animated (what objects can move and where they are in the image) and how to create an animation for their motion. In this work, we explore two frameworks for the generation task. In the first framework, we stack an autoencoder-like structure with a recurrent neural network layer to recursively generate future frames. The second framework adds a Deep Q-network to help the model make better decisions sequentially while generating future frames.

\subsection{Recurrent Model}
\label{sec:recurrent gen}

\begin{figure}[t]
\begin{center}
\includegraphics[width=0.9\linewidth]{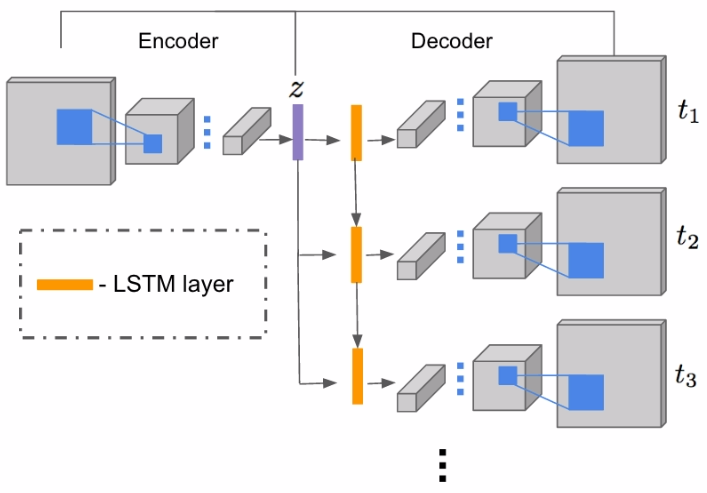}
\end{center}
   \caption{Recurrent generation architecture: Input to the network is an image, and the architecture consists of an autoencoder CNN stacked with an LSTM layer that recurrently generates future frames. }
\label{fig:rnn}
\end{figure}

In this scenario, we recursively generate frames of a cinemagraph, one by one, from a single static image. We do this by applying an autoencoder-style convolutional neural network stacked with an LSTM layer~\cite{LSTM1997}, recursively generating future frames based on a hidden variable $z$ (illustrated in Fig~\ref{fig:rnn}). An LSTM is ideal for our purpose due to its recurrent architecture and the ability to cope with temporal information of relatively long sequences.

Inspired by work on recursively generating future frames for timelapse videos of objects~\cite{temp}, we design our network with a similar structure and define a loss function with two tasks. The first is a mean square error (MSE) loss, that is, pixel-wise mean square error between generated outputs and the ground truth frames:
\begin{align}
  & loss_{mse}  = | \mathrm{Y} - G(x) |^2 \;&
\end{align}
where $x$ is the input image (the first frame in a cinemagraph sequence), $G(.)$ is the output of the generative model, and $\mathrm{Y}$ is the ground truth cinemagraph sequence.

The second part of the loss function is an adversarial loss. We define an autoencoder as the generator and introduce an additional binary CNN classifier as the discriminator. The discriminator classifies a given input image as real or fake (i.e. generated). These two components, generator and discriminator, can be seen as adversaries as they operate in competition with one another, and forms a generative adversarial network (GAN). Several previous generation works \cite{GANs,DCGAN,laplacian,play_plug,stackgan} have used GANs for generating high quality images. %The generator is trying to create images that are realistic enough to fool the discriminator into thinking they are real, while the discriminator is trying to differentiate between real images and those generated by the generator.  

\smallskip
\noindent The adversarial loss is represented as:
\begin{align}
  &loss_{adv}  = -\log(D[G(x)]) \;&
\end{align}
where D[.] is the output from the discriminator. This loss encourages the generated frames look like real images. Our dual-task loss function is then represented as 
\begin{align}
\label{eq3}
  &loss = loss_{mse} + \lambda_{adv} * loss_{adv} \;&
\end{align} 
where $\lambda_{adv}$ is a hyper-parameter that controls the impact of adversarial loss.

The recurrent model partially solve our sequential generation task. However, for cinemagraph generation, which is sightly different from video frame generation, we need a more expressive model with high stochasticity because the model needs to reason about the notion of foreground (moving) and background (fixed). Next, we discuss one possible way to achieve this by using Deep Q-network~\cite{ATARI}.

\subsection{Recurrent Deep Q-Network}
\label{sec:recurrent q-net}

\begin{figure}[t]
\begin{center}
\includegraphics[width=0.9\linewidth]{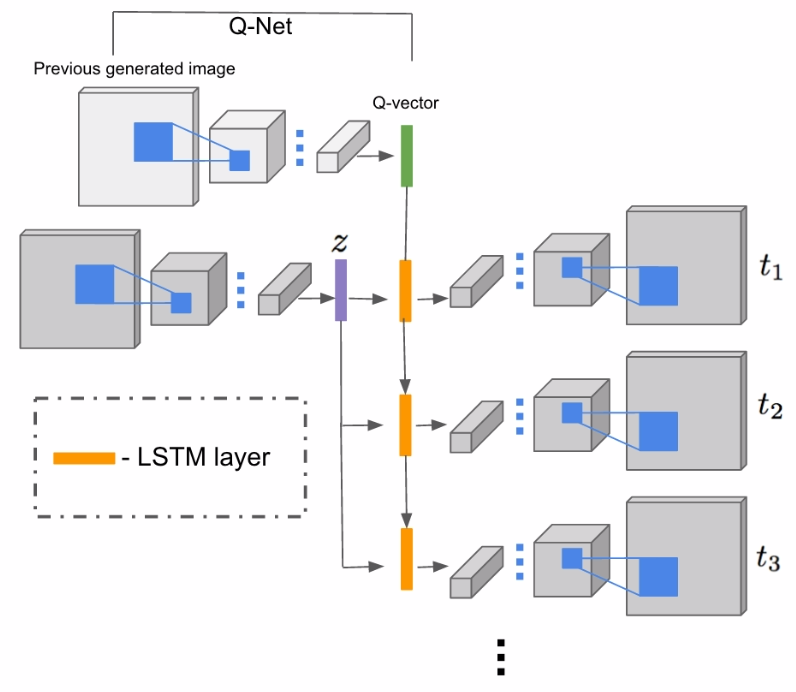}
\end{center}
   \caption{Recurrent generator with a deep Q-network. The Q-vector is concatenated with the latent state from an LSTM layer at each time step. Note that we show the concatenation only in the first time step to reduce clutter. }
\label{fig:rnn_q}
\end{figure}

Recent work~\cite{ATARI} has applied deep Q-learning to train a model to play Atari games, where the model makes decisions for each time step regarding how to adjust a joystick. Similarly, our cinemagraph generation task is a sequential decision making process. During generation, the model must decide which action to apply to generate the next frame. 

Inspired by the success of deep Q-learning, we incorporate a Deep Q-network (DQN), which is a convolutional neural network trained with Q-learning, into our framework. For DQN, the input is a state $s$ (e.g., game screenshots in Atari games) and the output is a Q-vector. The dimension of the Q-vector is equal to the number of possible actions (e.g. up and down). The value $q(s,a)$ for each dimension is defined as the discounted cumulative maximum future expected reward of taking action $a$ from state $s$. 
\begin{align}
  & q(s,a)  =  \underset{\pi}{\mathrm{max}}\mathbf{E} [r_t + \sigma r_{t+1} + \sigma ^2 r_{t+2} ... |s_t,a_t=s,a,\pi] \;&
\end{align} 
where $r_{t}$ is a reward at time step $t$, $\sigma$ is a discount factor, and $\pi$ is a policy mapping input states to actions. For each time step, the policy (what action to take when in state $s$) would then pick the action that maximizes the expected future reward in a greedy manner.

In this work, we apply DQN in a supervised generation task to make decisions about what ``action'' to take to generate the next frame. We first quantize a continuous action space into a discrete space (a Q-vector with limited dimensions). We then make use of the decoding structure of our generative model to decode how a ``quantized action'' will affect the state to generate the next frame. We evaluate how sensitive our model is to the size of the discretized space in Section~\ref{sec:abalation}
%\yale{This paragraph should go to related work.}
% One main difference between our task and Atari, is that in Atari the action space is discrete and low-dimensional (e.g. agents can only move up and down), while in our generation task, the action space is continuous. Additionally our action space is generally unnameable because an 'action' here is an abstract concept while in Atari the action space is easily defined (e.g. agents can move in certain directions from a state $s$). We also cannot directly obtain the next state by applying the 'action' to the current state as the next state is computed by applying a deep learning model. 
%To overcome these differences for our task,

Specifically, we propose a generation model cooperating with a DQN. The generator structure is the same as the recurrent model described in Sec.~\ref{sec:recurrent gen}. We introduce a DQN whose input is the previous generated frame from the generator (or the input image for the first time step). For the output, we quantize the continuous action space into a discrete space by defining the output Q-vector as an $N$ dimension vector to decide what ``action'' to take for a given state. We then encode it to a one-hot vector by assigning the maximal value of the Q-vector to 1 and others to 0. For each time step, we compute a different Q-vector based on the previous generation and concatenate it with the latent variable from the LSTM layer (as illustrated in Fig.~\ref{fig:rnn_q}). We let the concatenated vector pass through the decoding part to generate the frame showing the next state.

As we show in Sec.~\ref{evaluation}, adding the DQN to our recurrent generative model helps us achieve superior results -- in terms of both the sequence generation aspect (Sec.~\ref{real_results}) and the ``action'' prediction aspect, i.e., how the foreground object will move in the next step (Sec.\ref{sec:dqn_function}).

\begin{figure}[t]
\begin{center}
\includegraphics[width=1\linewidth]{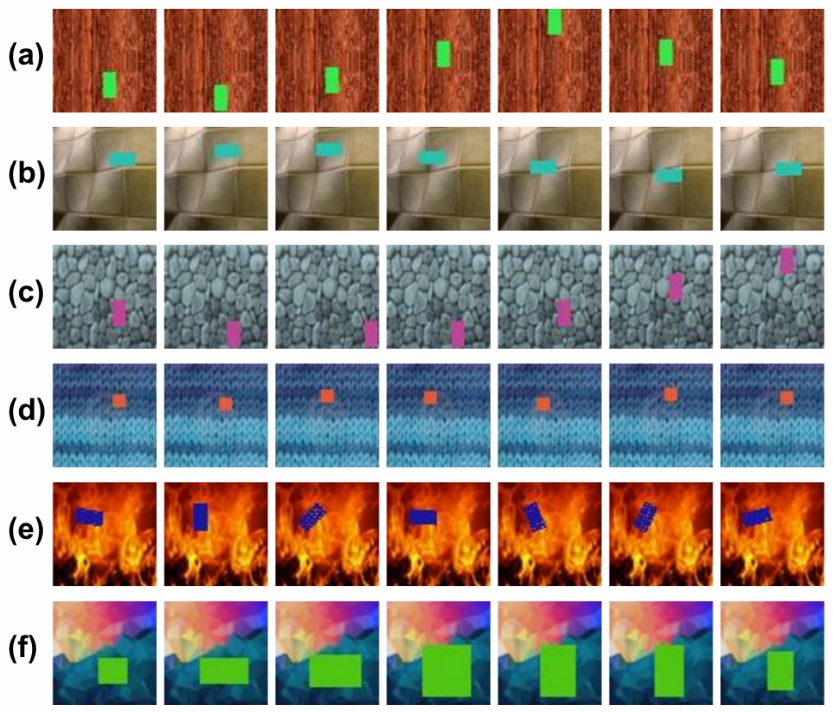}
\end{center}
   \caption{We show frames of six categories sub-sampled from synthetic generated sequences. (a) ``I'' Pattern frames; (b) ``O'' Pattern frames; (c) ``L'' Pattern frames; (d) ``Eight'' Pattern frames; (e) ``Rotate'' Pattern; and (f) ``Scale'' Pattern frames.}
\label{fig:sync_data}
\end{figure}

\begin{figure*}[t]
\begin{center}
\includegraphics[width=0.95\linewidth]{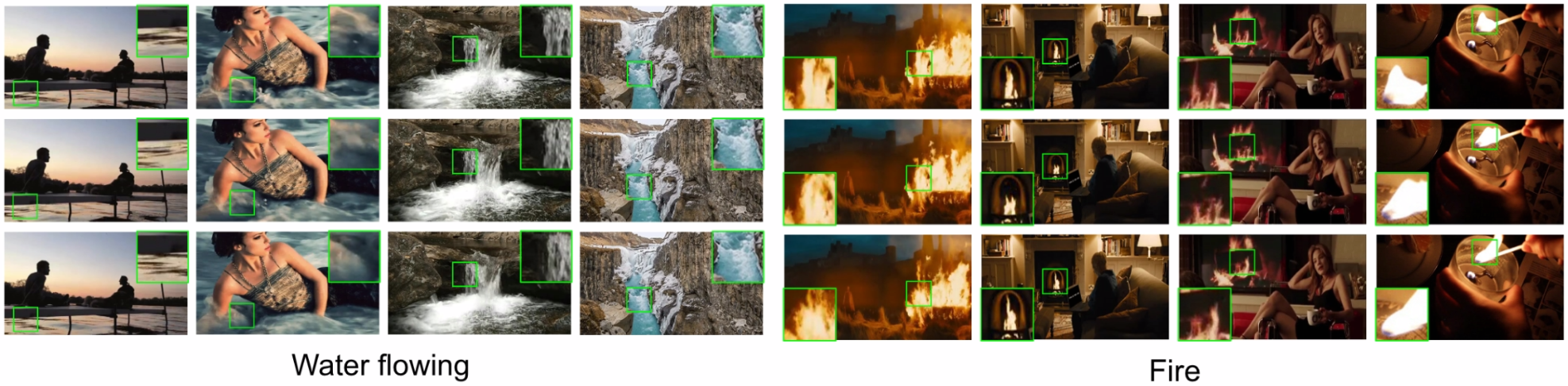}
\end{center}
   \caption{Example frames of ``Water flowing'' and ``fire'' categories from our real cinemagraph dataset. We show the cropped patches on the up-right or left-down corner to indicate the details. We observe along the columns that the corresponding objects are indeed moving.}
\label{fig:real_data}
\end{figure*}

%-------------------------------------------------------------------
\section{Datasets}
\label{datasets}
We collected two datasets: synthetic and real. The synthetic data allows us to evaluate our models with a large amount of data in a controlled setting, and to objectively measure performances with clearly defined ground truth labels. The real-data, on the other hand, allows us to test whether our models can indeed tackle the real-world problem of generating cinemagraphs from a single image.

\subsection{Synthetic Dataset}
To emulate a cinemagraph-like data, we collect 10 cluttered texture images and randomly pick multiple random offsets between [0.90, 1.1] for 3 channels as the fixed background. We then draw a randomly sized rectangle filled with a random color to one of the texture images as the ``foreground object.'' The foreground object will move in one of 6 moving patterns: ``I'' pattern, ``O'' pattern, ``L'' pattern, ``8'' pattern, rotate (counter clockwise), and scale (horizontally and vertically). Fig.~\ref{fig:sync_data} shows example frames from the synthetic sequences.
%1) 'I' Pattern: the object moves up and down.
%2) 'O' Pattern: the object moves in a circle motion.
%3) 'L' Pattern: the object moves up and down and from left to right like the letter L.
%4) 'Eight' Pattern: the object moves in a figure eight motion.
%5) 'Rotate' Pattern: the object rotates anti-clockwise.
%6) 'Scale' Pattern: the horizontal and vertical scale of the object changes over time. 

\begin{figure*}[t]
\begin{center}
\includegraphics[width=0.92\linewidth]{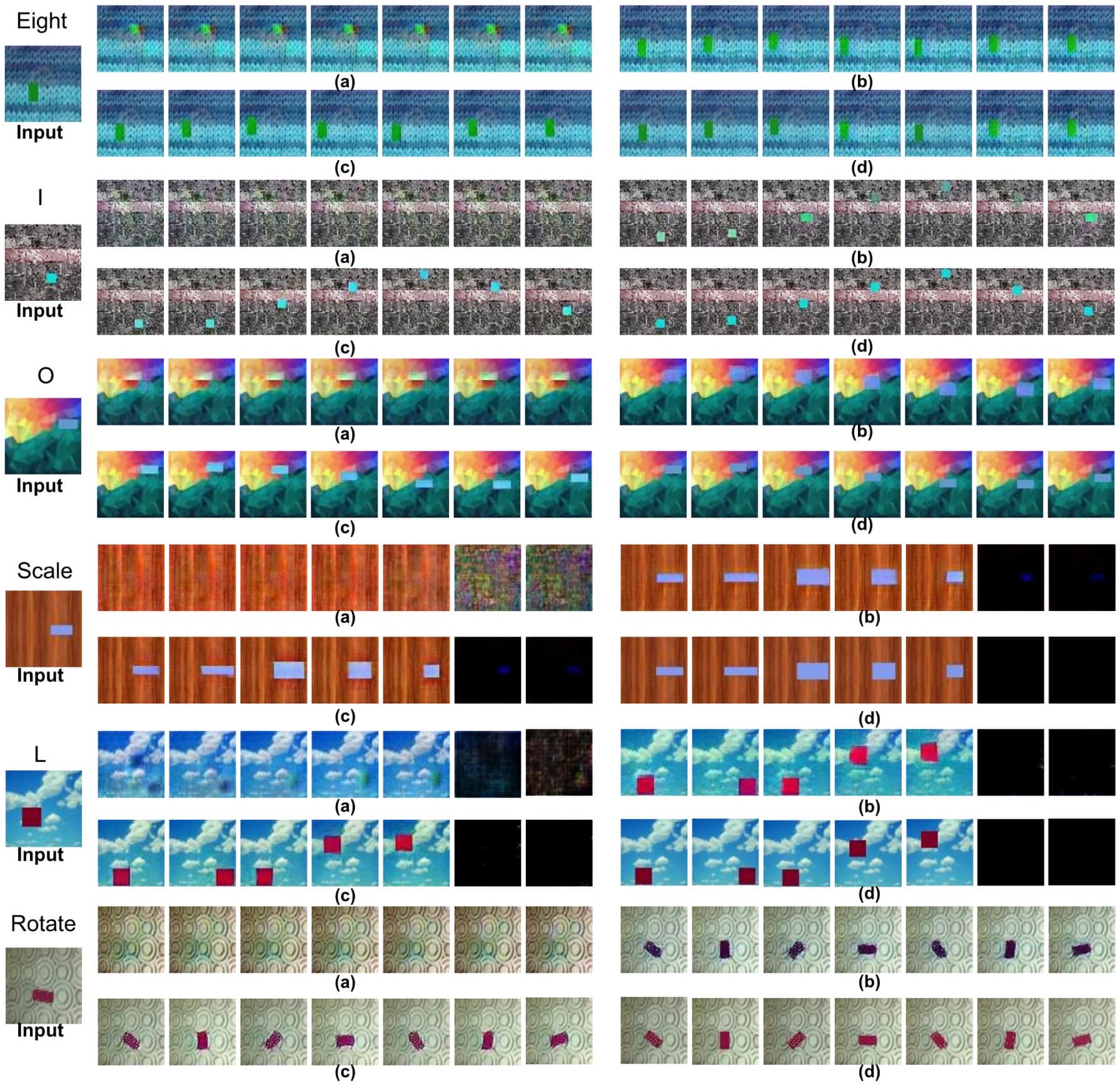}
\end{center}
   \caption{Qualitative results on synthetic data. We show an input frame along with results from a) Mathieu~\etal\cite{Yann}; (b) our RNN model; (c) our RNN+DQN model. (d) is the ground truth. We show 7 frames uniformly sampled from each sequence.}
\label{fig:sync_result}
\end{figure*}

\begin{table*} [t]
\begin{center}
\scalebox{0.8}{
\begin{tabular}{|l |c |c|| c| c || c| c|| c |c|| c| c|| c | c |}
\cline{2-13}
\multicolumn{1}{c|}{}&\multicolumn{2}{c||}{I}&\multicolumn{2}{c||}{O}&\multicolumn{2}{c||}{L}&\multicolumn{2}{c||}{Eight}&\multicolumn{2}{c||}{Rotate}&\multicolumn{2}{c|}{Scale}\\
\cline{2-13}
\multicolumn{1}{c|}{}& PSNR & SSIM & PSNR & SSIM  & PSNR & SSIM & PSNR& SSIM & PSNR & SSIM & PSNR & SSIM\\
\hline
Mathieu~\etal~\cite{Yann}  & 19.9801 &0.7917& 21.1132 & 0.8134& 9.9721 & 0.0631 & 19.9538 & 0.8004 & 20.7091&0.7820 & 14.4876 & 0.4502\\
RNN& 24.2417 & 0.8652& 26.4261 & 0.8879 & 27.2542 &0.8641& 26.5649 & 0.8861 & 27.2592&0.9101 &25.2220& \textbf{0.7611}\\
RNN+DQN & \textbf{25.4979} & \textbf{0.8672} & \textbf{27.9527} & \textbf{0.9210} & \textbf{27.3036} & \textbf{0.8782} & \textbf{27.5839} & \textbf{0.9161} & \textbf{27.3559}& \textbf{0.9115} & \textbf{25.5889}&0.7483\\
%{\bf0.8879}
\hline
\end{tabular}
}
\end{center}
\caption{PSNR and SSIM scores of Mathieu~\etal~\cite{Yann}, our RNN and RNN+DQN models on synthetic data across 6 categories.}
%\vspace{-.5cm}
\label{table:sync_result}
\end{table*}

\subsection{Cinemagraph Dataset}
To show the ability of our models generating cinemagraphs from natural images, we collect a dataset of real cinemagraphs from the Giphy website. We crawl GIFs with the tag ``cinemagraph'' and manually annotate the data with our predefined category names, e.g., water flowing, water pouring, fire, and candle light. We then select those categories with more than 200 samples. The resulting dataset contains 2 categories of cinemagraphs, depicting water flowing and fire. The number of cinemagraphs in the water flowing and fire categories are 926 and 350, respectively, with a total of 1,276 cinemagraphs. %We are currently collecting additional cinemagraphs from Giphy website where the data is growing quickly to obtain more data and categories.
% due to limited numbers of videos for other categories. We want that for each of the category, the number of cinemagraphs is at least 200, which might be enough to train computational models after data augmentation. 

A cinemagraph with long duration usually contains a great deal of movement replication. To alleviate this, we cut the cinemagraphs to be no more than 1 second. For long cinemagraphs that are more than 3 seconds long, we sample two to three 1-second clips with a maximum margin. This helps us augment the data for training, since the movements at different time steps may look different. We make sure to assign clips from the same cinemagraph to the same training/testing split. 

Some example frames are shown in Fig.\ref{fig:real_data}. We observe that frames contain both static and dynamic parts depicting water flowing and fire moving. Since the two moving types are difficult to see in still image frames, we crop and enlarge some key patches to show details of moving objects.\footnote{We include examples of animated cinemagraphs in the demo video in the project webpage.}
%\yale{This should be discussed in Introduction. You can do this even without referring to Figure 5.}

\begin{figure*}[t]
\begin{center}
\includegraphics[width=0.9\linewidth]{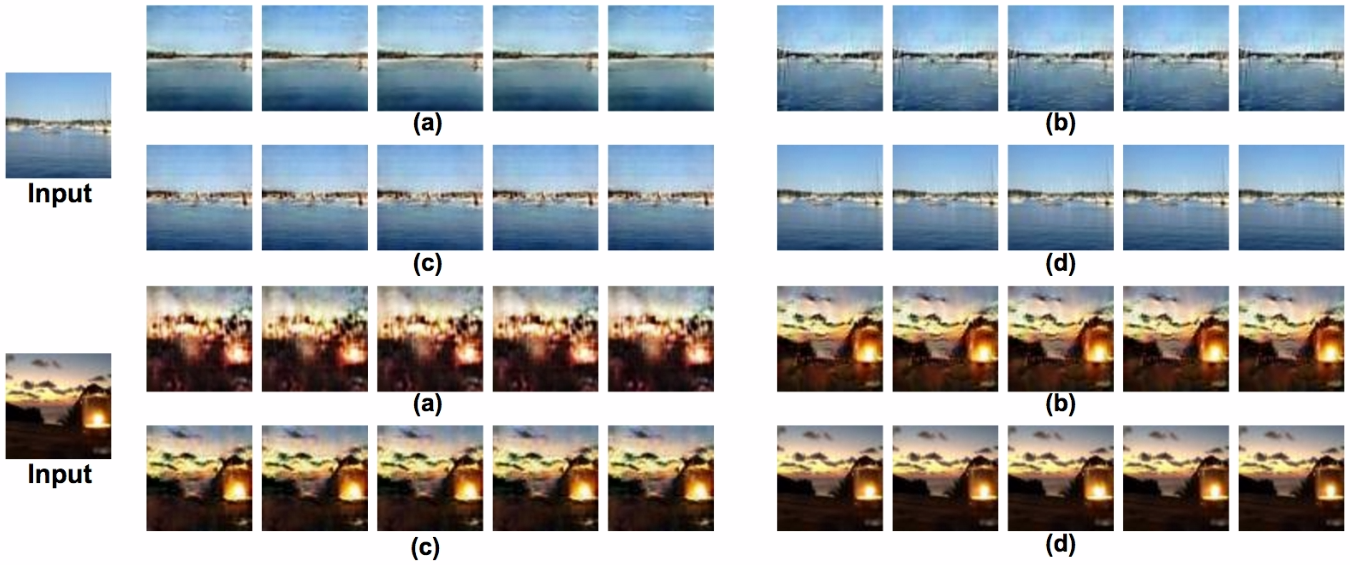}
\end{center}
   \caption{Qualitative results on the real cinemagraph data. For each sample, we show the input image and 5 generated frames generated by method: (a) Mathieu~\etal~\cite{Yann}; (b) RNN method; (c) RNN+DQN method; and (d) ground truth.}
\label{fig:real_result}
\end{figure*}

%-------------------------------------------------------------------
\section{Experiments}
\label{evaluation}

%-------------------------------------------------------------------
\subsection{Implementation Details}
\label{param}
For the generator of both the RNN (Sec \ref{sec:recurrent gen}) and the RNN+DQN (Sec \ref{sec:recurrent q-net}) models, inputs and outputs at each time step are of dimension 64x64x3, and the encoding and the decoding parts include 4 convolution/deconvolution layers with a kernel of size 5x5 and a stride of 2. The decoding part of each time step shares the same weights, and the number of feature maps for the layers are 64, 28, 256, 512/512, 256, 128, 64 respectively. The size of the hidden variable $z$ is 64 for synthetic data and 512 for real cinemagraph data generation. After each layer, we use ReLU as the activation function; for the last layer we apply a tanh activation function. 

For the RNN+DQN model, the architecture of DQN is the same as the encoding part of the generator, except we do not apply the tanh activation on the output layer. The output Q-vector is a 64 dimensional vector for synthetic data task, and 512 dimension for the real cinemagraph generation task.

We implement all models using Tensorflow. For all experiments, we use an ADAM optimizer~\cite{Adam} with a learning rate of 0.0002, and a batch size of 64. We set the weight for the adversarial loss $\lambda_{adv}$ to 0.005 for synthetic data and to 0.05 for real data experiments (equation \ref{eq3}). We raise the $\lambda_{adv}$ in real data experiments in order to enhance the realistic appearance of natural images.

%-------------------------------------------------------------------
\subsection{Synthetic Data Results}
\label{sync_results}
We evaluate the generation performance of the RNN model (Sec \ref{sec:recurrent gen}), RNN+DQN model (Sec \ref{sec:recurrent q-net}), and a baseline generation method from Mathieu~\etal~\cite{Yann} on 6 categories of synthetic data. For each category, we generate 100K sequences (95K for training and 5K for testing). Our proposed RNN and RNN+DQN models receive one input frame and recurrently generate the future frames. For each synthetic category, we train the model for 6K iterations. We note that Mathieu~\etal~\cite{Yann} is stateless, in that it takes the previous output as the input for the next time step to generate the sequence. For their model we experiment with several parameters settings, and find that applying the default setting described in their paper works the best. We also train their model for 6K iterations. 

For the synthetic data, we sample frames of sequences of each category. If the length of sampled sequences in one category is different, we pad the short sequences with a black image of the same size. The maximum length for 6 categories are: I Pattern : 27; O pattern : 21; L Pattern : 26; Eight Pattern : 21; Rotate pattern : 21;  Scale pattern : 17.

Fig.\ref{fig:sync_result} qualitative results. We observe that the baseline method~\cite{Yann} has a difficult time modeling the temporal movement of the foreground objects. We believe this is because it generates the future frame conditioned on only the previous (several) inputs while the sequence is relatively long. The superior quality of results of our RNN and RNN+DQN models suggest that they successfully learned the moving patterns of different categories. We can also see that the RNN+DQN model produces slightly more visually appealing results (the foreground objects have sharper boundary and more accurate shapes) than RNN method.

% And for 'L pattern' category which is the hardest case, after tune the parameters carefully we fail to train the model with the balanced generative and discriminative losses (yipin: still trying ...). 

We also quantitatively evaluate the methods. For the evaluation metrics, we compute the Peak Signal to Noise Ratio (PSNR)~\cite{psnr_ssim} and Structural Similarity Index (SSIM)~\cite{psnr_ssim} between a generated sequence and the ground truth sequence. Table~\ref{table:sync_result} shows that the RNN+DQN model works better than the RNN model. This is because, although both can generate a sharp background, RNN+DQN generates more accurate foreground objects. These two methods both significantly outperform the baseline method~\cite{Yann}, which does not apply an RNN structure to incorporate temporal information.

\subsection{Real Data Results}
\label{real_results}
We evaluate our RNN and RNN+DQN models, as well as a baseline~\cite{Yann} method, on ``water flowing'' and ``fire'' categories of our real cinemagraph dataset. For each cinemagraph, we sample 6 frames (1 input and 5 generation outputs). We split the training and testing sets with a ratio of 0.85 : 0.15. We augment the training data by cropping different regions from frames (random location fixed across time) and flipping frames left-right across time. For all three methods and categories, we train the models for 13,500 iterations. 

We first evaluate the method qualitatively. Fig.\ref{fig:real_result} shows input frames and generated frames under each method. We refer to our project website for detailed examples of generated cinemagraphs. The baseline method~\cite{Yann} achieves decent results because the length of the generation sequences is shorter than the synthetic data. However, similar to the case with synthetic data, background and foreground is still not well distinguished.

Overall, our models learn the correct motions for each object. The results suggest that our RNN and RNN+DQN models better differentiate and animate the foreground motions of water flowing and fire flickering, while keeping background fixed.\footnote{We note that, in the resulting frames shown in the paper, the foreground motion is hard to observe due to small image size and the nature of these small motions. We show videos of the generated cinemagraphs in our project webpage to better illustrate the results.} For the ``fire'' category, the resulting animated fires sometimes wave less strongly than the ground truth due to the limited data and high variance. 

For a quantitative evaluation, we compute PSNR and SSIM values between generated results and the ground truth sequences. Table~\ref{table:real_result} shows the results. RNN+DQN, which generates more visually appealing results, also achieves the best quantitative performance compared to the other two methods.

\begin{table} [t]
\begin{center}
\scalebox{0.95}{
\begin{tabular}{|l |c |c|| c| c |}
\cline{2-5}
\multicolumn{1}{c|}{}&\multicolumn{2}{c||}{Water flowing}&\multicolumn{2}{c|}{Fire}\\
\cline{2-5}
\multicolumn{1}{c|}{}& PSNR & SSIM & PSNR & SSIM \\
\hline
Mathieu~\cite{Yann}  & 18.2375 &0.4280& 18.2191 & 0.3784\\
RNN& 18.4333 & 0.4420& 19.4841 & 0.4831  \\
RNN+DQN &\textbf{19.4192} & \textbf{0.5053} & \textbf{20.2682} & \textbf{0.4976} \\
%{\bf0.8879}
\hline
\end{tabular}
}
\end{center}
\caption{Quantitative results (PSNR and SSIM scores) on the real cinemagraph dataset.}
%\vspace{-.5cm}
\label{table:real_result}
\end{table}

%%%%% from rebuttal %%%%%%%
\begin{table*} [t]
\begin{center}
\scalebox{0.6}{
\begin{tabular}{|l |c| c |  c| c| c| c|c ||c|c|c|}
\cline{2-11}
\multicolumn{1}{c|}{}&\multicolumn{1}{c|}{I}&\multicolumn{1}{c|}{O}&\multicolumn{1}{c|}{L}&\multicolumn{1}{c|}{Eight}&\multicolumn{1}{c|}{Rotate}&\multicolumn{1}{c|}{Scale}&\multicolumn{1}{c||}{Average}&\multicolumn{1}{c|}{Water}&\multicolumn{1}{c|}{Fire}&\multicolumn{1}{c|}{Average}\\
\cline{2-11}
\hline
Constant & 27.1174/0.5541& 27.9468/0.6270 & 26.7426/0.6097 & 28.9732/0.6361 & 29.4275/0.5893 & \textbf{31.0325/0.6459} & 28.5400/0.6104 & 29.3681/0.5768 & 28.7743/0.5434 & 29.0712/0.5601  \\
RNN+DQN & \textbf{27.9192/0.5576}  & \textbf{31.0523/0.6868} & \textbf{27.4687/0.6167} & \textbf{31.7454/0.6941} & \textbf{30.5351/0.6186} & 30.5754/0.6081 & \textbf{29.8827/0.6303} & \textbf{29.4949/0.5873} & \textbf{29.4030/0.5565} & \textbf{29.4490/0.5719}\\
%{\bf0.8879}
\hline
\end{tabular}
}
\end{center}
\caption{PSNR/SSIM scores of synthetic and real data and their averages on Constant baseline and RNN+DQN. (Col 2-8: synthetic data; Col 9-11: real data.)}
%\vspace{-.5cm}

\label{table:constant}
\end{table*}

\begin{table} [t]
\begin{center}
\scalebox{0.78}{
\begin{tabular}{|l |c| c |  c| c| c| c |}
\cline{2-7}
\multicolumn{1}{c|}{}&\multicolumn{1}{c|}{I}&\multicolumn{1}{c|}{O}&\multicolumn{1}{c|}{L}&\multicolumn{1}{c|}{Eight}&\multicolumn{1}{c|}{Rotate}&\multicolumn{1}{c|}{Scale}\\
\cline{2-7}
\hline
RNN& 2.9278  & 2.2839& 2.6734 & 2.4735 & 2.0150 &2.3675\\
RNN+DQN & \textbf{2.6440} & \textbf{2.1125} & \textbf{2.6285} & \textbf{2.3181} & \textbf{1.9931} & \textbf{2.1842}\\
%{\bf0.8879}
\hline
\end{tabular}
}
\end{center}
\caption{Average Euclidean distance between centers of the foreground object from the ground truth and from generated frames.}
%\vspace{-.5cm}

\label{table:dist}
\end{table}

\begin{table} [t]
\begin{center}
\scalebox{0.9}{
\begin{tabular}{|l |c| c |}
\cline{2-3}
\multicolumn{1}{c|}{}&\multicolumn{1}{c|}{Synthetic}&\multicolumn{1}{c|}{Real}\\
\cline{2-3}
\hline
Original  & \textbf{26.8805 / 0.8737}  & \textbf{19.8437 / 0.5015}\\
Small $z$& 21.3788 / 0.7794  & 13.7626 / 0.1364  \\
Large $z$ & 26.2362 / 0.8670 & 18.4141 / 0.4111 \\
Small Q-vec& 25.8371 / 0.8556  & 17.9049 / 0.3974 \\
Large Q-vec & 26.1532 / 0.8585 & 19.6085 / 0.4885 \\
%{\bf0.8879}
\hline
\end{tabular}
}
\end{center}
\caption{The average PSNR/SSIM scores with different dimensions of $z$ and Q vectors. }
%\vspace{-.5cm}
\label{table:ablation}
\end{table}

\begin{table} [t]
\begin{center}
\scalebox{0.8}{
\begin{tabular}{|l |c |c || c | c|}
\cline{2-5}
\multicolumn{1}{c|}{}&\multicolumn{1}{c|}{Water flowing}&\multicolumn{1}{c||}{Fire}&\multicolumn{1}{c|}{Water flowing}&\multicolumn{1}{c|}{Fire}\\
\cline{2-5}
\hline
Mathieu~\cite{Yann}  & 30.00\%  & 26.42\% & 11.43\%  & 20.75\%\\
RNN&  32.86\%  & 35.84\%  & 27.86\% & 26.41\% \\
RNN+DQN & \textbf{37.14\%} & \textbf{37.74\%} & \textbf{35.00\%} & \textbf{33.96\%} \\
%{\bf0.8879}
\hline
\end{tabular}
}
\end{center}
\caption{Human evaluation results. Column 2-3 shows how often humans preferred one method over the others. Column 4-5 shows how often humans were ``fooled'' into believing that the generated results are real cinemagraphs.}
%\vspace{-.5cm}
\label{table:human}
\end{table}

%-------------------------------------------------------------------
\subsection{Constant Baseline}
To further show that the proposed method is able to capture the (foreground) moving objects while making the background static, we provide a simple ``constant'' baseline, i.e., new frames are simply copied from the first frame. To capture the temporal dynamics, we compute PSNR/SSIM on frame-wise difference images (two consecutive frames) rather than on the original frames. For the constant baseline, frame-wise difference results in all-zero images because there is no movement over time. 
Table~\ref{table:constant} shows that RNN+DQN method achieves better performance than the baseline, which suggests the capability of our method discriminating foreground objects from backgrounds.

%In this session, we only use input images as a constant evaluation baseline. For both synthetic and real data, we show the PSNR/SSIM evaluation in Table~\ref{table:constant}. The PSNR/SSIM values of constant baseline of synthetic data is much lower than both RNN and RNN+DQL methods while for the real data the situations are the opposite due to the blurriness of generated images. But still our proposed algorithms can successfully discriminate foreground and background and generate reasonable cinemagraphs. More than 30\% of generated cinemagraphs from the proposed methods can fool humans (Sec \ref{sec: human_exp}). (maybe combine with last two sessions, and unfinished yet.)

\subsection{Deep Q-Network}
\label{sec:dqn_function}
Deep Q-network (DQN) helps improve the results by bringing in more stochasticity into our model, improving sequential decision making and thus the overall visual quality of generated cinemagraphs. Here, we show the ability of DQN to help decide what objects in a scene should move and how they should move. 

We conduct an experiment on synthetic dataset to evaluate the model's ability to predict the location of moving objects (find which object to move and how to move). Specifically, we pre-compute the segmentation maps (segmenting the moving object and the background) of the synthetic data. During testing, we segment out the object from the ground truth data and run a sliding window on the generated frames to get the resulting segmentation maps. And we compute the average Euclidean distance between the object centers of ground truth and generated frames. 

Table~\ref{table:dist} shows the results. The results suggest that adding DQN indeed improves the prediction quality on how objects move in future frames. We do not evaluate Mathieu~\etal~\cite{Yann} because it often fails to generate moving objects, and thus the sliding window technique cannot be applied. 

\subsection{Sensitivity Analysis}
\label{sec:abalation}
We conduct an ablation study to show how sensitive the results are to the size of $z$ (bottleneck of the auto-encoder) and the Q-vector (output of DQN). Specifically, we evaluate the RNN+DQN model with small (4-dim) and large (2048-dim) sizes of $z$ and Q-vector, and compare them with the original results we presented earlier (64 for synthetic data and 512 for real data)

Table~\ref{table:ablation} reports average PSNR/SSIM scores of small and large $z$/Q-vector on both synthetic and real datasets. We notice that utilizing the small bottleneck and/or Q-vector significantly harm the results due to the severe information loss and over quantization (except for applying small Q-vector on synthetic dataset because the action space of partial synthetic categories is relatively small). %While still competitive, the results are slightly worse than the original. We conjecture this is because of the information redundancy and also the GPU memory increases significantly.

\subsection{Human Evaluation}
\label{sec: human_exp}
There is a universal problem in evaluating video prediction methods: There might be multiple correct predicted results. There might be multiple cinemagraphs that share the first frame but look different from each other, all of which could be considered reasonable to a human observer. Therefore, we design two human experiments to judge the quality of real cinemagraph generated by different methods. 

In the first experiment, we show human subjects three cinemagraphs generated from different methods (RNN, RNN+DQN, Mathieu~\etal~\cite{Yann}) and ask them to choose one that looks the most superior. We used Amazon Mechanical Turk (Mturk) for this study. Results are shown in Table~\ref{table:human} (Col 2-3), where numbers indicate how often humans selected results from each method as the best cinemagraph. We note that RNN+DQN method achieves the best results under both categories.

In the second experiment, we show human subjects a single cinemagraph (either real or generated) at a time and ask them whether it looks realistic. We measure how often humans choose the generated ones as real and report the results in Table~\ref{table:human} (Col 4-5), which shows that our RNN+DQN is the most successful at this task.

\section{Conclusion}
In this paper, we explore a challenging task of generating a cinemagraph from a single image. We propose a method that combines recurrent generative models with a deep Q-network to learn what regions should move and how they should move over time. On both synthetic and real datasets, we evaluate qualitative and quantitative results of the proposed methods and show improved performance over a baseline method of Mathieu~\etal~\cite{Yann}. 

Future work includes expanding the natural cinemagraph dataset to additional categories with increased numbers of examples. We also plan to analyze the learned models to better understand what the models are learning about image content semantics and motions of objects in the world.

%-------------------------------------------------------------------

{\small
\bibliographystyle{ieee}
\bibliography{egbib}

\begin{thebibliography}{10}\itemsep=-1pt

\bibitem{Wasserstein_GAN}
M.~Arjovsky, S.~Chintala, and L.~Bottou.
\newblock Wasserstein generative adversarial networks.
\newblock In {\em ICML}, 2017.

\bibitem{BEGAN}
D.~Berthelot, T.~Schumm, and L.~Metz.
\newblock {BEGAN:} boundary equilibrium generative adversarial networks.
\newblock {\em CoRR}, 2017.

\bibitem{laplacian}
E.~L. Denton, S.~Chintala, A.~Szlam, and R.~Fergus.
\newblock Deep generative image models using a laplacian pyramid of adversarial
  networks.
\newblock In {\em NIPS}. 2015.

\bibitem{Unsuper_physical}
C.~Finn, I.~J. Goodfellow, and S.~Levine.
\newblock Unsupervised learning for physical interaction through video
  prediction.
\newblock In {\em NIPS}, 2016.

\bibitem{GANs}
I.~Goodfellow, J.~Pouget-Abadie, M.~Mirza, B.~Xu, D.~Warde-Farley, S.~Ozair,
  A.~Courville, and Y.~Bengio.
\newblock Generative adversarial nets.
\newblock In {\em NIPS}. 2014.

\bibitem{DRAW}
K.~Gregor, I.~Danihelka, A.~Graves, D.~Rezende, and D.~Wierstra.
\newblock Draw: A recurrent neural network for image generation.
\newblock In {\em ICML}, 2015.

\bibitem{q_continue}
S.~Gu, T.~Lillicrap, I.~Sutskever, and S.~Levine.
\newblock {Continuous Deep Q-Learning with Model-based Acceleration}.
\newblock {\em ICML}, 2016.

\bibitem{LSTM1997}
S.~Hochreiter and J.~Schmidhuber.
\newblock Long short-term memory.
\newblock {\em Neural Comput.}, 1997.

\bibitem{pix2pix2016}
P.~Isola, J.-Y. Zhu, T.~Zhou, and A.~A. Efros.
\newblock Image-to-image translation with conditional adversarial networks.
\newblock {\em CVPR}, 2017.

\bibitem{VPN}
N.~{Kalchbrenner}, A.~{van den Oord}, K.~{Simonyan}, I.~{Danihelka},
  O.~{Vinyals}, A.~{Graves}, and K.~{Kavukcuoglu}.
\newblock {Video Pixel Networks}.
\newblock {\em CoRR}, 2016.

\bibitem{progressive_GAN}
T.~Karras, T.~Aila, S.~Laine, and J.~Lehtinen.
\newblock Progressive growing of gans for improved quality, stability, and
  variation.
\newblock {\em CoRR}, 2017.

\bibitem{helicopter}
H.~J. Kim, M.~I. Jordan, S.~Sastry, and A.~Y. Ng.
\newblock Autonomous helicopter flight via reinforcement learning.
\newblock In {\em NIPS}. 2004.

\bibitem{Adam}
D.~P. Kingma and J.~Ba.
\newblock {Adam: {A} Method for Stochastic Optimization}.
\newblock {\em ICLR}, 2014.

\bibitem{trajectory}
T.~Kollar and N.~Roy.
\newblock Trajectory optimization using reinforcement learning for map
  exploration.
\newblock {\em International Journal of Robotics Research}, 2008.

\bibitem{Liang_2017_ICCV}
X.~Liang, L.~Lee, W.~Dai, and E.~P. Xing.
\newblock Dual motion gan for future-flow embedded video prediction.
\newblock In {\em ICCV}, 2017.

\bibitem{continuous}
T.~P. Lillicrap, J.~J. Hunt, A.~Pritzel, N.~Heess, T.~Erez, Y.~Tassa,
  D.~Silver, and D.~Wierstra.
\newblock Continuous control with deep reinforcement learning.
\newblock In {\em ICLR}, 2016.

\bibitem{Yann}
M.~Mathieu, C.~Couprie, and Y.~LeCun.
\newblock Deep multi-scale video prediction beyond mean square error.
\newblock {\em CoRR}, 2015.

\bibitem{obstacle}
J.~Michels, A.~Saxena, and A.~Y. Ng.
\newblock High speed obstacle avoidance using monocular vision and
  reinforcement learning.
\newblock In {\em ICML}, 2005.

\bibitem{Asy}
V.~Mnih, A.~P. Badia, A.~G. Mehdi~Mirza, T.~P. Lillicrap, T.~Harley, D.~Silver,
  and K.~Kavukcuoglu.
\newblock Asynchronous methods for deep reinforcement learning.
\newblock In {\em ICML}, 2016.

\bibitem{ATARI}
V.~Mnih, K.~Kavukcuoglu, D.~Silver, A.~A. Rusu, J.~Veness, M.~G. Bellemare,
  A.~Graves, M.~Riedmiller, A.~K. Fidjeland, G.~Ostrovski, S.~Petersen,
  C.~Beattie, A.~Sadik, I.~Antonoglou, H.~King, D.~Kumaran, D.~Wierstra,
  S.~Legg, and D.~Hassabis.
\newblock Human-level control through deep reinforcement learning.
\newblock {\em Nature}, 2015.

\bibitem{play_plug}
A.~Nguyen, J.~Yosinski, Y.~Bengio, A.~Dosovitskiy, and J.~Clune.
\newblock Plug {\&} play generative networks: Conditional iterative generation
  of images in latent space.
\newblock {\em CVPR}, 2017.

\bibitem{action_atari}
J.~Oh, X.~Guo, H.~Lee, R.~L. Lewis, and S.~Singh.
\newblock Action-conditional video prediction using deep networks in atari
  games.
\newblock In {\em NIPS}. 2015.

\bibitem{CVPR16context}
D.~Pathak, P.~Kr\"ahenb\"uhl, J.~Donahue, T.~Darrell, and A.~Efros.
\newblock Context encoders: Feature learning by inpainting.
\newblock In {\em CVPR}, 2016.

\bibitem{DCGAN}
A.~Radford, L.~Metz, and S.~Chintala.
\newblock Unsupervised representation learning with deep convolutional
  generative adversarial networks.
\newblock {\em CoRR}, 2015.

\bibitem{Ranzato2014}
M.~Ranzato, A.~Szlam, J.~Bruna, M.~Mathieu, R.~Collobert, and S.~Chopra.
\newblock Video (language) modeling: a baseline for generative models of
  natural videos.
\newblock {\em CoRR}, 2014.

\bibitem{SSIM}
K.~Ridgeway, J.~Snell, B.~Roads, R.~S. Zemel, and M.~C. Mozer.
\newblock Learning to generate images with perceptual similarity metrics.
\newblock {\em CoRR}, 2015.

\bibitem{highdim}
J.~Schulman, P.~Moritz, S.~Levine, M.~Jordan, and P.~Abbeel.
\newblock {High-Dimensional Continuous Control Using Generalized Advantage
  Estimation}.
\newblock {\em ICLR}, 2016.

\bibitem{deter}
D.~Silver, G.~Lever, N.~Heess, T.~Degris, D.~Wierstra, and M.~Riedmiller.
\newblock Deterministic policy gradient algorithms.
\newblock In {\em ICML}, 2014.

\bibitem{LSTM}
N.~Srivastava, E.~Mansimov, and R.~Salakhudinov.
\newblock Unsupervised learning of video representations using lstms.
\newblock In {\em ICML}, 2015.

\bibitem{Hierarchical_future}
R.~Villegas, J.~Yang, Y.~Zou, S.~Sohn, X.~Lin, and H.~Lee.
\newblock Learning to generate long-term future via hierarchical prediction.
\newblock {\em ICML}, 2017.

\bibitem{scene_dynamics}
C.~Vondrick, H.~Pirsiavash, and A.~Torralba.
\newblock Generating videos with scene dynamics.
\newblock In {\em NIPS}, 2016.

\bibitem{jacob_iccv2017}
J.~Walker, K.~Marino, A.~Gupta, and M.~Hebert.
\newblock The pose knows: Video forecasting by generating pose futures.
\newblock In {\em ICCV}, 2017.

\bibitem{Style_gen}
X.~Wang and A.~Gupta.
\newblock Generative image modeling using style and structure adversarial
  networks.
\newblock In {\em ECCV}, 2016.

\bibitem{psnr_ssim}
Z.~Wang, A.~C. Bovik, H.~R. Sheikh, and E.~P. Simoncelli.
\newblock Image quality assessment: from error visibility to structural
  similarity.
\newblock {\em IEEE Transactions on Image Processing}, 2004.

\bibitem{visual_dynamics}
T.~Xue, J.~Wu, K.~L. Bouman, and W.~T. Freeman.
\newblock Visual dynamics: Probabilistic future frame synthesis via cross
  convolutional networks.
\newblock In {\em NIPS}, 2016.

\bibitem{stackgan}
H.~Zhang, T.~Xu, H.~Li, S.~Zhang, X.~Huang, X.~Wang, and D.~N. Metaxas.
\newblock Stackgan: Text to photo-realistic image synthesis with stacked
  generative adversarial networks.
\newblock {\em ICCV}, 2017.

\bibitem{energy}
J.~{Zhao}, M.~{Mathieu}, and Y.~{LeCun}.
\newblock {Energy-based Generative Adversarial Network}.
\newblock {\em CoRR}, 2016.

\bibitem{temp}
Y.~Zhou and T.~L. Berg.
\newblock Learning temporal transformations from time-lapse videos.
\newblock In {\em ECCV}, 2016.

\bibitem{CycleGAN2017}
J.-Y. Zhu, T.~Park, P.~Isola, and A.~A. Efros.
\newblock Unpaired image-to-image translation using cycle-consistent
  adversarial networks.
\newblock {\em ICCV}, 2017.

\bibitem{navi}
Y.~Zhu, R.~Mottaghi, E.~Kolve, J.~J. Lim, L.~F.-F. Abhinav~Gupta, and
  A.~Farhadi.
\newblock {Target-driven Visual Navigation in Indoor Scenes using Deep
  Reinforcement Learning}.
\newblock {\em ICRA}, 2017.

\end{thebibliography}
}

\end{document}